\documentclass{article}
\usepackage{ijcai26}

\usepackage{times}
\usepackage{soul}
\usepackage{url}
\usepackage{amssymb}
\usepackage[hidelinks]{hyperref}
\usepackage[utf8]{inputenc}
\usepackage[small]{caption}
\usepackage{graphicx}
\usepackage{amsmath}
\usepackage{amsthm}
\usepackage{booktabs}
\usepackage{algorithm}
\usepackage{algorithmic}
\usepackage[switch]{lineno}
\usepackage{xcolor}

\usepackage{enumitem}
\usepackage{bm}
\usepackage{pbalance}
\newcommand{\sstitle}[1]{\noindent{\bf #1\/.}}

\theoremstyle{definition}
\newtheorem{definition}{Definition}

\title{Prior Knowledge-enhanced Spatio-temporal Epidemic Forecasting
}

\author{
Sijie Ruan$^1$\and
Jinyu Li$^1$\and
Jia Wei$^1$\and
Zenghao Xu$^2$\and
Jie Bao$^3$\and
Junshi Xu$^4$\textnormal{,}\\
Junyang Qiu$^5$\and
Shuliang Wang$^1$\and
Xiaoxiao Wang$^2$\thanks{Xiaoxiao Wang and Hanning Yuan are corresponding authors.} ~\textnormal{and}
Hanning Yuan$^{1\ast}$
\affiliations
$^1$Beijing Institute of Technology, Beijing, China \quad \\
$^2$Zhejiang Provincial Center for Disease Control and Prevention, Hangzhou, China \\
$^3$JD Technology, Beijing, China \\
$^4$The University of Hong Kong, Hong Kong SAR, China \\
$^5$China Mobile Internet, Guangzhou, China
\emails
\{sjruan, ljy666, weijia05, slwang2011, yhn6\}@bit.edu.cn,
\{zhxu, xxwang\}@cdc.zj.cn,\\
baojie@jd.com,
junshixu@hku.hk,
qiujunyang@139.com
}

\begin{document}

\maketitle

\begin{abstract}
Spatio-temporal epidemic forecasting is critical for public health management, yet existing methods often struggle with insensitivity to weak epidemic signals, over-simplified spatial relations, and unstable parameter estimation. To address these challenges, we propose the \underline{S}patio-\underline{T}emporal pri\underline{O}r-aware \underline{E}pidemic \underline{P}redictor (STOEP), a novel hybrid framework that integrates implicit spatio-temporal priors and explicit expert priors. STOEP consists of three key components: (1) Case-aware Adjacency Learning (CAL), which dynamically adjusts mobility-based regional dependencies using historical infection patterns; (2) Space-informed Parameter Estimating (SPE), which employs learnable spatial priors to amplify weak epidemic signals; and (3) Filter-based Mechanistic Forecasting (FMF), which uses an expert-guided adaptive thresholding strategy to regularize epidemic parameters. Extensive experiments on real-world COVID-19 and influenza datasets demonstrate that STOEP outperforms the best baseline by 11.1\% in RMSE. The system has been deployed at a provincial CDC in China to facilitate downstream applications.

\end{abstract}

\section{Introduction}

Population-level epidemic forecasting is of great importance to public health management, since accurate forecasting of new infections at each location can help governments and healthcare providers allocate resources, design intervention policies, and evaluate the effectiveness of control measures~\cite{qian2021cpas}. 

In the early years, constrained by data scarcity, epidemic forecasting primarily relied on mechanistic models, e.g., SIR~\cite{kermack1927contribution}. With the rise of deep learning, data-driven approaches have been increasingly adopted in this field~\cite{arora2020prediction}. Simultaneously, realizing the spatial dependencies of disease transmission among regions, spatio-temporal forecasting across multiple regions jointly has garnered great attention~\cite{deng2020cola}.

However, due to the limited epidemiological data, these models remain prone to overfitting. In recent years, there has been a surge focused on hybrid models that integrate mechanistic principles with deep learning~\cite{cao2022mepognn,wang2022causalgnn,han2025epidemiology}, e.g., employing deep neural networks to estimate the parameters of mechanistic models as shown in Figure~\ref{fig:intro}(a). Nevertheless, existing methods still encounter three primary limitations:

\begin{figure}[t]
  \begin{minipage}{0.48\textwidth}
    \includegraphics[width=\linewidth]{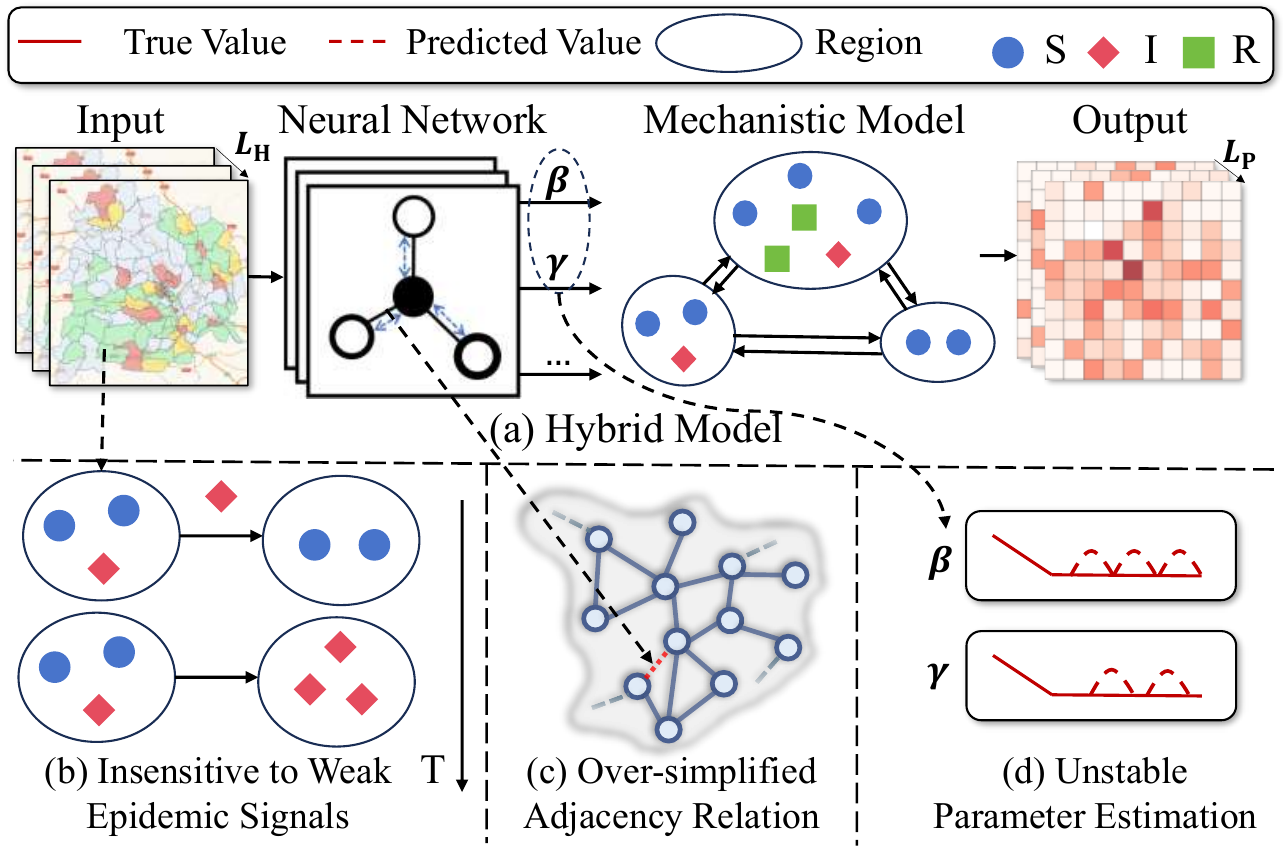}
    \caption{(a) Illustrations of hybrid models; (b) Insensitivity to weak epidemic signals; (c) Over-simplified adjacency relation; (d) Unstable parameter estimation.}
    \label{fig:intro}
  \end{minipage}
  \hfill
  \begin{minipage}{0.48\textwidth}~\end{minipage}
\end{figure}

\begin{itemize}[leftmargin=*]
\item \textbf{Insensitivity to Weak Epidemic Signals.} Unlike traffic forecasting, epidemic data often contain weak signals as shown in Figure~\ref{fig:intro}(b). That is, the confirmed cases in most of the time are small, but infectious diseases can surge to high prevalence levels in a short time. Existing deep-learning modules~\cite{wu2019graph} are often insensitive to such weak yet critical signals, leading to missed early warnings.
    \item \textbf{Over-simplified Adjacency Relation.} 
    Existing methods capture the spatial dependency by using the human mobility intensities among regions~\cite{wang2018inferring,cao2022mepognn}. Though the vital factor, i.e., human mobility for disease transmission, is considered, they omit the intrinsic similarities among regions as shown in Figure~\ref{fig:intro}(c).
    \item \textbf{Unstable Parameter Estimation.}
    As mentioned, hybrid models rely on neural networks to produce vital epidemic parameters.
    However, unconstrained network outputs often fluctuate violently in data-scarce scenarios as shown in Figure~\ref{fig:intro}(d), resulting in mechanically implausible and unstable parameter estimation.
\end{itemize}

An intuitive idea to tackle those issues is to incorporate the external epidemic knowledge to regulate the modeling process. However, related data may not always be available. To this end, in this paper, we propose \underline{S}patio-\underline{T}emporal pri\underline{O}r-aware \underline{E}pidemic \underline{P}redictor (STOEP), which incorporates two types of implicit knowledge without the requirements of external data, i.e., spatio-temporal priors and expert priors, to facilitate the epidemic modeling. The spatio-temporal priors are derived from data to enhance the representation of each region even if the signals are weak, and adjust the adjacency relation among regions in addition to the mobility data, while the expert priors are obtained from domain experts to regularize the output of estimated epidemic parameters to stabilize the epidemic forecasting. More specifically, STOEP is composed of three modules: 
(1) \textit{Case-aware Adjacency Learning (CAL)}, which dynamically fuses mobility data with patterns of confirmed cases by a learnable pattern memory to capture temporal epidemic priors dynamically;
(2) \textit{Space-informed Parameter Estimating (SPE)}, which introduces a specialized spatial enhancement block with learnable spatial priors to explicitly amplify signals;
(3) \textit{Filter-based Mechanistic Forecasting (FMF)}, which employs an adaptive thresholding strategy guided by expert priors to suppress noise in data-scarce regions.
The main contributions are summarized as follows:

\begin{itemize}[leftmargin=*]
    \item We inject spatio-temporal priors into the epidemic forecasting model, which makes the adjacency relation aware of temporal patterns via CAL, and makes model inputs enhanced by spatial knowledge via SPE.
    \item We propose an expert prior-guided adaptive thresholding strategy via FMF to stabilize the epidemic forecasting by regularizing the output of estimated epidemic parameters. 
    \item Extensive experiments on two real-world epidemic datasets, i.e., COVID-19 and influenza, demonstrate that STOEP outperforms the best baseline by 11.1\% on average in RMSE. We also released the code for public use\footnote{https://github.com/stdi-lab/STOEP}.
    \item A system based on STOEP is deployed and used at a provincial CDC in China to facilitate downstream applications.
\end{itemize}

\section{Preliminaries} 

\subsection{Definitions}

\begin{definition}[Daily Confirmed Cases]
We denote the number of daily confirmed cases in region $n$ at day $\tau$ as $x^{\tau}_n$, and the number of daily confirmed cases over all regions as $\mathbf{X}^{\tau}\in \mathbb{R}^{N}$, where $N$ is the number of regions.
\end{definition}

\begin{definition}[Historical Observations]
To make the epidemic forecasting model work effectively, we divide historical observations into two types, i.e., essential factors and optional factors. The essential factors include historical daily confirmed cases $\mathbf{X}^{t-(T_{\text{in}}-1):t}$, susceptible population $\mathbf{S}^{t-(T_{\text{in}}-1):t} \in \mathbb{R}^{N \times T_{\text{in}}}$, infected population $\mathbf{I}^{t-(T_{\text{in}}-1):t} \in \mathbb{R}^{N \times T_{\text{in}}}$ and recovered population $\mathbf{R}^{t-(T_{\text{in}}-1):t} \in \mathbb{R}^{N \times T_{\text{in}}}$, where $T_{in}$ is the given lookback time window. Other related features can also be incorporated as optional factors, e.g., day of week.
We denote historical observations as $\mathcal{O}^{t-(T_{\text{in}}-1):t} \in \mathbb{R}^{N \times T_{\text{in}} \times C}$, where $C\ge 4$ is the number of factors considered.
\end{definition}

\begin{definition}[Historical Mobility Data]
We use historical mobility data
$\mathcal{M}^{\,t-(T_{\text{in}}-1):t} \in \mathbb{R}^{N \times N \times T_{\text{in}}}$ to characterize the mobility intensity among different regions, where entry $m_{ij}^{\tau}\!\ge 0$ denotes the flow intensity from region $i$ to region $j$ at time $\tau \in \{t-(T_{\text{in}}-1),\ldots,t\}$.
\end{definition}

\subsection{Problem Statement}
Given the historical observations $\mathcal{O}^{t-(T_{\text{in}}-1):t} \in \mathbb{R}^{N \times T_{\text{in}} \times C}$ and mobility data
$\mathcal{M}^{\,t-(T_{\text{in}}-1):t} \in \mathbb{R}^{N \times N \times T_{\text{in}}}$, predict the daily confirmed cases in the future $T_{\text{out}}$ timesteps, i.e., $\hat{\mathbf{X}}^{t+1:t+T_{out}}\in \mathbb{R}^{N\times T_{out}}$, which can be formulated as:

\begin{equation}
    \{\mathcal{O}^{t-(T_{\text{in}}-1):t}, \mathcal{M}^{\,t-(T_{\text{in}}-1):t}\} \xrightarrow{f(\cdot)} \hat{\mathbf{X}}^{t+1:t+T_{out}}
\end{equation}

\section{Methodology}

\begin{figure*}[htbp]
  \centering
  \includegraphics[width=\textwidth]{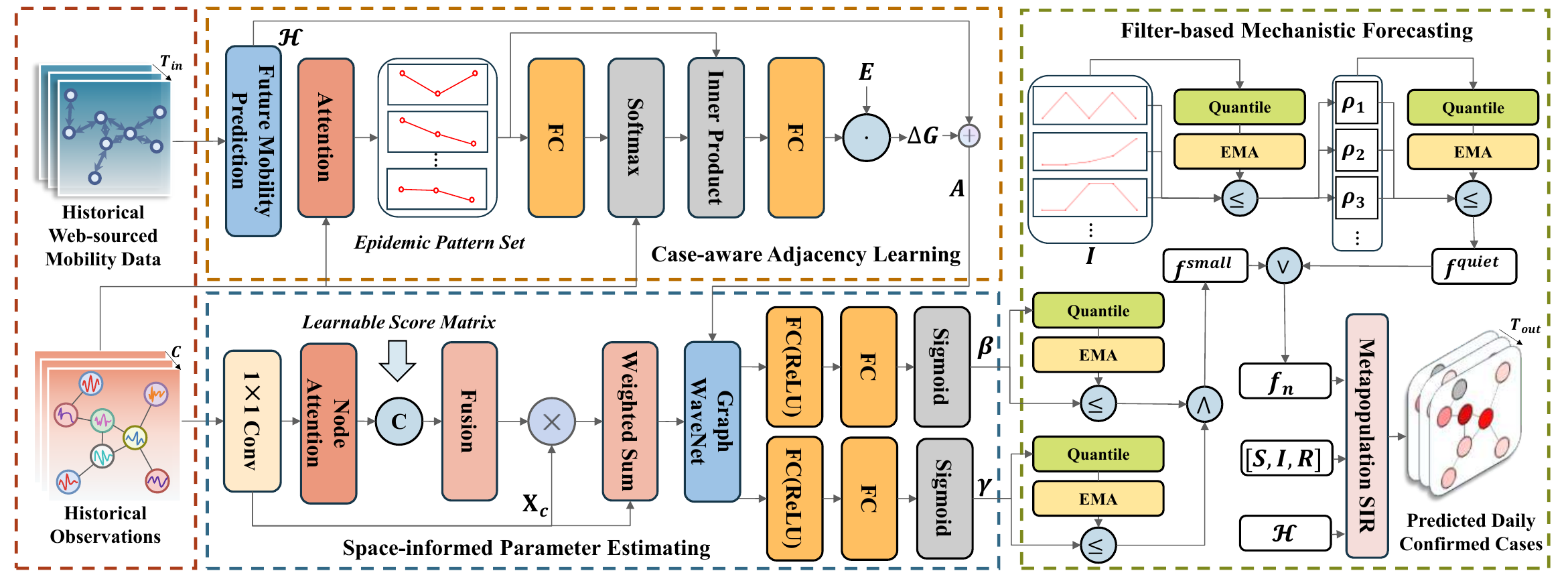}
  \caption{Framework of proposed epidemic forecasting model STOEP.}
  \label{fig:framework}
\end{figure*}

STOEP follows the paradigm of hybrid spatio-temporal epidemic forecasting, which essentially leverages a mechanistic model to forecast future confirmed cases, but the epidemic parameters of the model are estimated by neural networks. The architecture of STOEP is depicted in Figure~\ref{fig:framework}, which consists of three modules:

\begin{itemize}[leftmargin=*]
    \item \textbf{Case-aware Adjacency Learning (CAL)}, which takes historical mobility data as well as daily confirmed cases, and produces a case-aware adjacency matrix. It would be used to indicate the adjacency intensities. 
    \item \textbf{Space-informed Parameter Estimating (SPE)}, which takes historical observations, learned spatial priors and case-aware adjacency matrix, to estimate the epidemic parameters of the mechanistic model.
    \item \textbf{Filter-based Mechanistic Forecasting (FMF)}, which takes estimated epidemic parameters as well as the human mobility data, predicts the future daily confirmed cases via a filter-based mechanistic model.
\end{itemize}

The complexity and scalability of STOEP is practical for real-world use, and the detailed analysis is given in Appendix~\ref{sec:appendix_complexity}.
Next, we elaborate on each module in detail. 

\subsection{Case-aware Adjacency Learning}
Case-aware adjacency learning (CAL) aims to learn the case-aware adjacency intensities among different regions $\mathbf{A}\in \mathbb{R}^{N\times N}$ based on the historical mobility data $\mathcal{M}^{\,t-(T_{\text{in}}-1):t}$ and daily confirmed cases $\mathbf{X}^{t-(T_{\text{in}}-1):t}$. 

\sstitle{Main Idea} As mentioned, to adequately model the adjacency relation among regions, we fuse the mobility-based intensities with data-derived intensities. 
The main idea is to adjust the mobility-based intensities using data-derived adjacency relationship, which are detailed as follows. 

\sstitle{Mobility-based Adjacency Learning} We first perform a linear transformation from past to future on $\mathcal{M}$ to obtain the future mobility prediction $\mathcal {H}^{t+1:t+T_{\text{out}}}\in \mathbb{R}^{N\times N\times T_{\text{out}}}$.
And the mobility-based adjacency matrix $\mathbf{A}_m\in \mathbb{R}^{N\times N}$ can be derived by mean-pooling over the prediction horizons:

\begin{equation}
\mathbf{A}_m \;=\; \frac{1}{T_{\text{out}}}\sum_{i=1}^{T_{\text{out}}} \mathcal H^{\,t+i} \label{eq:dynamicA}
\end{equation}

\sstitle{Case-aware Adjacency Adjustment} For each region, its recent observation of confirmed cases may contain some noise. To quantify its correlation with other regions based on the recently observed cases, we first perform a soft clustering to obtain its robust representation, and then use the representation to calculate the intensity with other regions.

More specifically, for a historical daily confirmed case of region $n$, i.e., $\mathbf{x}^{t-(T_{\text{in}}-1):t}_n$, we take its most recent $S$ days ($S\le T_{\text{in}}$) to form a recent observation $\mathbf{x}^{t-(S-1):t}_n$. $\mathbf{x}^{t-(S-1):t}_n$ would be normalized to eliminate the effect of population disparity to form an epidemic pattern $\tilde{\mathbf{x}}^{t-(S-1):t}_n$. 
Then the robust representation $\mathbf{r}_n\in \mathbb{R}^{d_{A}}$ of the current epidemic pattern of region $n$ can be derived by attention-based pattern retrieval from a learnable pattern set $\mathbf{P}\in \mathbb{R}^{P\times S}$:

\begin{equation}
\bm\alpha_n \;=\; \mathrm{softmax}\!\left(\frac{\bm q_n \mathbf K^\top}{\sqrt d}\right), \qquad
\mathbf{r}_n \;=\; \text{FC}(\bm\alpha_n \mathbf V) \label{eq:attn}
\end{equation}
where $\bm\alpha_n$ are the attention weights of region $n$ over $P$ patterns,  
$\mathbf{K}\in \mathbb{R}^{d}$ and $\mathbf{V}\in \mathbb{R}^{d}$ are transformed key and value matrices of $\mathbf{P}$.
After that, $\mathbf{r}_n$ would be compared with vectors in a learnable region embedding table $\mathbf{E}\in \mathbb{R}^{N\times d_A}$ to obtain its recent correlation with other regions. Formally, the case-based adjacency intensity $\Delta G_{nm}$ of region $n$ to region $m$ is:  

\begin{equation}
    \Delta G_{nm} \;=\; \alpha \cdot \langle \bm r_n,\, \bm e_m \rangle \label{eq:edge}
\end{equation}
where $\bm e_m\in\mathbb{R}^{d_A}$ is entry $m$ of $\mathbf E$,
$\langle\cdot,\cdot\rangle$ denotes the inner product. $\alpha$ is a learnable scalar that scales the correction strength, which is initialized as 0 to prevent training perturbations.
Finally, the case-based adjacency $\Delta \mathbf{G}\in \mathbb{R}^{N\times N}$ serves as a residual term to adjust the mobility-based adjacency $\mathbf{A}_m$ to obtain the final adjacency matrix $\mathbf{A}$:

\begin{equation}
    \mathbf A \;=\; \mathbf{A}_m \;+\; \Delta \mathbf{G}\label{eq:dynamicAA}
\end{equation}

\subsection{Space-informed Parameter Estimating}

Space-informed parameter estimating (SPE) aims to learn epidemic parameters of the mechanistic model. 
In this work, we leverage MetaSIR~\cite{wang2018inferring} as the mechanistic model, since it can jointly model multiple regions. The basic principle of MetaSIR can be found in Appendix~\ref{sec:metasir}.
In the context of MetaSIR, STOEP needs to estimate the infection rate $\bm{\beta}\in \mathbb{R}^{N\times T_{\text{out}}}$ and recovery rate $\bm{\gamma}\in \mathbb{R}^{N\times T_{\text{out}}}$ of different regions in future time steps. 

\sstitle{Main Idea}
To amplify the weak signals in epidemic data, we aim to explicitly learn some stable spatial dependency, which would be fused with dynamic dependency derived from historical observations, to comprehensively enhance the input features.
After that, the enhanced features are fed into GraphWaveNet~\cite{wu2019graph} to estimate $\bm{\beta}$ and $\bm{\gamma}$, which are detailed as follows.

\sstitle{Dynamic Dependency Generation}
We first transform the historical observations $\mathcal{O}^{t-(T_{\text{in}}-1):t}$ to a hidden representation $\mathbf{X}_c \in\mathbb{R}^{N\times T_{\text{in}}\times C_c}$ by a $1\times 1$ convolution. We compute the inter-region dynamic dependency attention map $\mathbf{A}^{\text{node}}\in \mathbb{R}^{N\times N}$ by applying multi-head self-attention to $\mathbf{X}_c$ for every input step, and averaging the scores over time and heads:

\begin{align}
\mathbf A^{\text{node}} = \frac{1}{H T_\text{in}} \sum_{h,t} \text{Attn}_\text{h}(\mathbf X_{c,t}^{(h)}).
\end{align}
where \(H\) is the number of heads.

\sstitle{Spatial Prior Incorporation}
To capture stable and long-range spatial dependency, we introduce a learnable score matrix $\mathbf{S} \in \mathbb{R}^{N \times N}$ that encodes a global spatial prior. $\mathbf{S}$ is initialized as an identity matrix and optimized during the training. We apply a row-wise softmax on $\mathbf{S}$ to obtain a static dependency attention map $\mathbf{A}^{\text{struct}}=\operatorname{softmax}_{\text{row}}(\mathbf{S})$.

We use a gating mechanism to fuse the dynamic dependency and the static dependency to obtain a fused attention map $\mathbf{A}^{\text{fuse}}\in \mathbb{R}^{N\times N}$ as follows:
\begin{equation}
\begin{aligned}
\mathbf \Gamma \;&=\; \phi\!\big([\mathbf A^{\text{node}};\,\mathbf A^{\text{struct}}]\big)\\
\mathbf A^{\text{fuse}} \;&=\; \sigma(\mathbf \Gamma)\odot \mathbf A^{\text{node}} + \big(1-\sigma(\mathbf \Gamma)\big)\odot \mathbf A^{\text{struct}} \label{eq:fuse}
\end{aligned}
\end{equation}
where \([\cdot;\cdot]\) is the concatenation, \(\phi(\cdot)\) is a linear layer to produce a gate \(\mathbf \Gamma\in\mathbb{R}^{N\times N}\), $\sigma(\cdot)$ is sigmoid activation, and \(\odot\) is Hadamard product.
After that, we regularize $\mathbf{A}^{\text{fuse}}$ by symmetrizing, enforcing nonnegativity, and row-normalizing to obtain a refined attention map $\mathbf{A}^{\text{fuse}}_{\text{norm}}\in \mathbb{R}^{N\times N}$.

\sstitle{Feature Enhancement}
Based on $\mathbf{A}^{\text{fuse}}_{\text{norm}}$, we enhance the features $\mathbf{X}_c$ in a residual connection manner:
\begin{equation}
\begin{aligned}
\bar {\mathbf X}_{c} \;&=\; \mathbf A^{\text{fuse}}_{\text{norm}}\, \mathbf X_{c}\\
\tilde{\mathbf{O}}^{t-(T_{\text{in}}-1):t} \;&=\; (1-\epsilon)\,\mathbf X_c \;+\; \epsilon\,\bar {\mathbf X}_c
\end{aligned}
\end{equation}
where $\tilde{\mathbf{O}}^{t-(T_{\text{in}}-1):t}\in\mathbb{R}^{N\times T_{\text{in}}\times C_c}$ is the feature enhanced input, and $\epsilon\in[0,1]$ is a learnable scalar (initialized as \(0\)).

\sstitle{Parameter Estimation}
After node features are enhanced, we feed $\tilde{\mathbf{O}}^{t-(T_{\text{in}}-1):t}$ as well as the adjacency matrix $\mathbf{A}$ obtained in CAL into GraphWaveNet-style backbone~\cite{wu2019graph} for spatio-temporal modeling to obtain the hidden representation $\mathbf{Z} \in\mathbb{R}^{N\times T_{\text{out}}\times d}$:

\begin{equation}
    \mathbf{Z} \;=\; \text{GraphWaveNet}(\tilde{\mathbf{O}}^{t-(T_{\text{in}}-1):t},\, \mathbf{A})
\end{equation}
Finally, two linear heads take the hidden representation and predict the epidemiological parameters $\bm{\beta}\in \mathbb{R}^{N\times T_{\text{out}}}$ and $\bm{\gamma}\in \mathbb{R}^{N\times T_{\text{out}}}$ for each node on future time steps:
\begin{equation}
    \bm{\beta} = \sigma(\text{FC}_\beta(\mathbf{Z})), \qquad \bm{\gamma} = \sigma(\text{FC}_\gamma(\mathbf{Z}))
\end{equation}

\subsection{Filter-based Mechanistic Forecasting}

Filter-based mechanistic forecasting (FMF) aims to forecast future daily confirmed cases $\hat{\mathbf{X}}^{t+1:t+T_{\text{out}}}$ based on the estimated epidemic parameters $\bm{\beta}$ and $\bm{\gamma}$ from SPE via a filter-based MetaSIR.

\sstitle{Main Idea}
To mitigate the instability of epidemic parameter estimation, by discussing with the domain experts, we suppress the disease transmission process if estimated epidemic parameters or infected population are small. A straightforward idea is to set a constant threshold to suppress the epidemic parameters. However, due to the heterogeneity across space and time, we propose an adaptive thresholding strategy to flexibly decide whether the disease transmission process should be suppressed or not, which are detailed as follows.

\sstitle{Adaptive Thresholding Strategy}
Adaptive thresholding strategy aims to learn an adaptive threshold given a set of numbers, e.g., $\{v_r\}_{r=1}^{R}$. We first compute the quantile $\kappa=\operatorname{Quantile}\big(\{v_r\}_{r=1}^{R},\,\kappa\big)$, where $\kappa$ is the desired quantile ratio. 
During the training, $\kappa$ would be smoothed by Exponential Moving Average (EMA) to control the adaptivity–stability trade-off.
The adaptive threshold is defined as 
\begin{equation}
\delta \;=\; \max\!\big\{\kappa,\; \delta_{\min}\big\}, \label{eq:threshold_batch}
\end{equation}
where $\delta_{\min}$ is a safeguard lower bound to avoid vanishing thresholds when the quantile becomes very small.

\sstitle{Small Parameter Detection}  
Small parameter detection aims to detect whether the estimated epidemic parameters $\bm{\beta}_n\in \mathbb{R}^{T_\text{out}}$ and $\bm{\gamma}_n\in \mathbb{R}^{T_\text{out}}$ of a region $n$ are sufficiently small. For each region $n$, we first obtain its maximum parameter estimates over the prediction horizon, i.e., $\beta_n^{max}$ and $\gamma_n^{max}$.
We then apply the adaptive thresholding over the maximum parameter set of all regions, i.e., $\{\beta_n^{max}\}_{N}$ and $\{\gamma_n^{max}\}_N$ with quantile $\kappa_\beta$ and $\kappa_\gamma$, to obtain adaptive thresholds $\beta_{\text{thr}}$ and $\gamma_{\text{thr}}$. A region with both small $\beta_n^{max}$ and $\gamma_n^{max}$ would be detected: 

\begin{equation}
    f_n^{\text{small}} = \mathbb{I}(\beta_n^{max} \le \beta_{\text{thr}} \land \gamma_n^{max} \le \gamma_{\text{thr}})
\end{equation}

\sstitle{Low Infection Detection}  
Low infection detection aims to detect whether the infected population of a region is sufficiently small. We first apply the adaptive thresholding on $\mathbf{I}^{t-(T_{\text{in}}-1):t}$ with quantile $\kappa_I$ to obtain a global near-zero threshold $I_{\text{abs}}$. Then, for each region $n$, the proportion of near-zero infection days can be defined as $\rho_n = \frac{1}{T_{\text{in}}} \sum_{\tau=t-(T_{\text{in}-1})}^{t} \mathbb{I}(I_n^{\tau} \le I_{\text{abs}})$.
We then apply the adaptive thresholding over $\{\rho_n\}_N$ with quantile $\kappa_\rho$ to obtain $\widehat{\kappa}_{\text{ZR}}$. To maintain suppression sensitivity, we cap the threshold by $\kappa_\rho^{\max}$, i.e., $\widehat{\kappa}_{\text{ZR}} \leftarrow \min\!\big(\widehat{\kappa}_{\text{ZR}},\, \kappa_\rho^{\max}\big)$.
A region with proportion of near-zero infection days larger than $\widehat{\kappa}_{\text{ZR}}$ would be detected:

\begin{equation}
f_n^{\text{quiet}} = \mathbb{I}(\rho_n \ge \widehat{\kappa}_{\text{ZR}}), \label{eq:quietmask}
\end{equation}

\sstitle{Suppressed Metapopulation SIR} 
The final suppression filter is defined as the union of two detection results, i.e., $f_n = f_n^{\text{quiet}} \lor f_n^{\text{small}}$.
We apply the filter to adaptively suppress 
the disease transmission process by modifying the infection parameter $\bm{\beta}_n$ in the MetaSIR model~\cite{wang2018inferring}, which is achieved as follows:

\begin{equation}
    \hat{\bm{\beta}}_n = \psi^{f_n}\, \bm{\beta}_n
\end{equation}
where $\psi \in (0,1)$ is a small constant for downscaling.
Based on the revised infection update equation, we can obtain the daily confirmed cases prediction according to MetaSIR~\cite{wang2018inferring}.
During training, the model is optimized via Mean Absolute Error (MAE) loss.

\section{Experiments}

We evaluate the performance of the proposed method on two real-world epidemic datasets.

\subsection{Datasets}

To validate the applicability of the proposed model, we conducted evaluation on two types of epidemic datasets with unique characteristics, which are summarized as follows:

\begin{table*}[t]
\centering
\fontsize{7pt}{8pt}\selectfont
\setlength{\tabcolsep}{3.5pt}
\renewcommand{\arraystretch}{1.15}
\begin{tabular}{lcccccccccccccccc}
\toprule
{\textbf{Model}} 
& \multicolumn{4}{c}{\textbf{3 d Ahead}} 
& \multicolumn{4}{c}{\textbf{7 d Ahead}} 
& \multicolumn{4}{c}{\textbf{14 d Ahead}} 
& \multicolumn{4}{c}{\textbf{Overall}} \\
\cmidrule(lr){2-5}\cmidrule(lr){6-9}\cmidrule(lr){10-13}\cmidrule(lr){14-17}
& RMSE & MAE & SMAPE & RAE
& RMSE & MAE & SMAPE & RAE
& RMSE & MAE & SMAPE & RAE
& RMSE & MAE & SMAPE & RAE \\
\midrule
SIR [1927]       & 441.0 & 156.5 & 84.1 & 0.49
                 & 522.3 & 195.2 & 113.3 & 0.58
                 & 932.1 & 322.9 & 133.2 & 0.97
                 & 616.8 & 214.6 & 102.7 & 0.65 \\
MetaSIR [KDD18]  & 346.8 & 128.4 & 72.7 & 0.39
                 & 412.4 & 167.8 & 94.8 & 0.51
                 & 795.4 & 283.2 & 129.7 & 0.84
                 & 517.4 & 184.3 & 110.4 & 0.54 \\ \hline
STGCN [IJCAI18]   & 372.9 & 117.0 & 44.9 & 0.36
                  & 378.5 & 127.0 & 52.1 & 0.38
                  & 427.9 & 158.5 & 74.2 & 0.48
                  & 388.4 & 131.6 & 55.3 & 0.40 \\
MTGNN [KDD20]    & 307.2 & 105.8 & 41.1 & 0.32
                 & 382.4 & 137.5 & 50.1 & 0.41
                 & 443.5 & 168.3 & 80.2 & 0.52
                 & 363.2 & 135.0 & 50.7 & 0.49 \\
CovidGNN [2020]   & 288.4 & 96.8 & 45.1 & 0.29
                  & 339.1 & 127.9 & 63.4 & 0.38
                  & 443.3 & 185.0 & 114.4 & 0.55
                  & 358.2 & 133.4 & 68.4 & 0.40 \\
DCRNN [ICLR18]   & 312.7 & 92.8 & 37.9 & 0.28
                  & 336.3 & 111.8 & 48.4 & 0.34
                  & 386.6 & 149.9 & 72.6 & 0.45
                  & 344.3 & 116.0 & 50.8 & 0.35 \\
PopNet [WWW22]   & 330.4 & 128.1 & 57.6 & 0.39
                 & 331.9 & 130.6 & 56.8 & 0.39
                 & 398.3 & 157.7 & 67.2 & 0.48
                 & 341.3 & 133.0 & 58.5 & 0.40 \\
TimeKAN [ICLR25] & 327.4 & 124.8 & 55.8 & 0.38
                 & 330.8 & 129.4 & 58.5 & 0.39
                 & 329.8 & 130.0 & 61.7 & 0.39
                 & 329.1 & 127.8 & 57.6 & 0.39 \\
COVID-Forecaster [ICDE23]& 318.6 & 106.7 & 50.0 & 0.32
                 & 322.1 & 118.7 & 58.8 & 0.35
                 & 331.9 & 140.9 & 80.1 & 0.42
                 & 324.0 & 121.4 & 62.2 & 0.37 \\
GraphWaveNet [IJCAI19]& 234.2 & 76.5  & 47.3 & 0.22
                 & 315.1 & 105.5 & 57.7 & 0.32
                 & 308.4 & 148.1 & 76.6 & 0.44
                 & 315.1 & 105.5 & 57.7 & 0.32 \\
LightGTS [ICML25]& 254.7 & 88.4 & 39.1 & 0.27
                 & 284.0 & 104.9 & 48.0 & 0.31
                 & 379.8 & 149.3 & 72.1 & 0.45
                 & 306.2 & 111.0 & 50.4 & 0.33 \\
ColaGNN [CIKM20] & 242.0 & 76.3  & 40.2 & 0.23
                 & 337.4 & 119.9 & 43.9 & 0.37
                  & 376.1 & 147.5 & 69.9 & 0.43
                  & 294.2 & 106.4 & 50.2 & 0.33 \\
AMD [AAAI25]       & 287.5 & 113.0 & 65.0 & 0.34
                 & 291.0 & 115.5 & 65.3 & 0.34
                 & 291.7 & 115.3 & 63.7 & 0.34
                 & 289.8 & 114.7 & 65.2 & 0.34 \\
STDDE [WWW24]    & 277.5 & 105.2 & 51.9 & 0.32
                 & 277.0 & 107.2 & 51.6 & 0.32
                 & 276.9 & 108.2 & 53.1 & 0.33
                 & 277.1 & 107.0 & 52.1 & 0.32 \\
PDFormer [AAAI23]& 273.1 & 97.7 & \underline{36.6} & 0.29
                 & 269.2 & 97.8 & 44.2 & 0.29
                 & 275.5 & 103.6 & 62.5 & 0.31
                 & 272.2 & 99.5 & \underline{46.3} & 0.30 \\
DUET [KDD25]     & 275.6 & 100.8 & 43.2 & 0.30
                 & 271.3 & 101.2 & \underline{44.0} & 0.30
                 & 267.6 & \underline{101.9} & 63.0 & \underline{0.30}
                 & 271.6 & 101.3 & 46.5 & 0.30 \\
AGCRN [NeurIPS20]& 209.1 & 76.1 & 49.9 & 0.23
                  & 234.0 & 94.1 & 55.7 & 0.28
                  & 336.9 & 143.2 & 81.5 & 0.43
                  & 253.5 & 103.0 & 60.0 & 0.31 \\ \hline
EISTGNN [EAAI25] & 352.5 & 132.6 & 61.3 & 0.40
                 & 350.7 & 135.2 & 62.2 & 0.40
                 & 416.0 & 159.6 & 68.7 & 0.48
                 & 369.3 & 141.0 & 63.7 & 0.42 \\
STAN [JAMIA21]   & 330.6 & 127.8 & 57.0 & 0.39
                 & 365.4 & 143.9 & 62.8 & 0.43
                 & 398.9 & 158.6 & 70.1 & 0.48
                 & 348.4 & 136.8 & 60.3 & 0.41 \\
CausalGNN [AAAI22] & 308.8 & 104.4 & 39.6 & 0.32
                 & 313.6 & 116.8 & 46.4 & 0.35
                 & 331.9 & 149.2 & 67.1 & 0.45
                 & 317.1 & 121.4 & 49.9 & 0.37 \\
MPSTAN [Entropy24] & 302.1 & 127.0 & 51.4 & 0.38
                 & 316.8 & 137.5 & 54.3 & 0.41
                 & 331.9 & 147.4 & \textbf{60.6} & 0.44
                 & 316.1 & 136.7 & 54.9 & 0.41 \\
MepoGNN [ECML-PKDD22] & \underline{139.2} & \underline{54.9}   & 37.6 & \underline{0.17}
                 & \underline{177.5}  & \underline{72.8}   & 45.5 & \underline{0.22}
                 & \underline{250.6}  & 103.8  & \underline{61.1} & 0.31
                 & \underline{188.5}  & \underline{75.5}   & 47.1 & \underline{0.23} \\ \hline\hline
\textbf{STOEP (Ours)}   & \textbf{125.3} & \textbf{49.8} & \textbf{35.9} & \textbf{0.15}
                & \textbf{149.3} & \textbf{63.1} & \textbf{43.1} & \textbf{0.19}
                & \textbf{230.1} & \textbf{97.9} & 61.6 & \textbf{0.29} 
                & \textbf{169.1} & \textbf{68.8} & \textbf{45.4} & \textbf{0.21} \\
\bottomrule
\end{tabular}
\caption{Results on the \textbf{COVID-19} dataset across different horizons (STOEP's improvements are statistically significant, with $p$-value $< 2 \times 10^{-4}$ across all baselines).}
\label{tab:JPepi_all_horizons}
\end{table*}

\begin{itemize}[leftmargin=*]
\item \textbf{COVID-19 Dataset.}  
This dataset contains daily confirmed COVID-19 cases in 47 prefectures in Japan covering the period from April 1, 2020, to September 21, 2021 (539 days) from the NHK COVID-19\footnote{https://www3.nhk.or.jp/news/special/coronavirus/data/}. Other related observations are from the Ministry of Health, Labour and Welfare, etc. The mobility data are from Facebook Movement Range Maps\footnote{https://data.humdata.org/dataset/movement-range-maps}. The entire dataset is publicly available\footnote{https://github.com/deepkashiwa20/MepoGNN}. 
\item \textbf{Flu Dataset.}  
This dataset contains daily confirmed flu cases in 11 cities of Zhejiang Province, China, during 2023-2024 (731 days).
We can derive the daily susceptible/infected/recovered population from it in a way similar to that used for the COVID-19 dataset.
The mobility data are from Baidu Migration Map\footnote{https://qianxi.baidu.com/}. 
\end{itemize}

The detailed demographic and mobility characteristics of both datasets can be found in Appendix~\ref{sec:appendix_dataset_analysis}.

For each dataset, we partition it into non-overlapping training, validation, and test sets chronologically at a 6:1:1 ratio.

% baselines
\subsection{Experimental Settings}
\sstitle{Baselines} 
We compare STOEP with the following baselines:
\begin{itemize}[leftmargin=*]
    \item \textbf{Mechanistic Models}: SIR~\cite{kermack1927contribution} and MetaSIR~\cite{wang2018inferring}.
    \item \textbf{Deep Learning Models}: 
        (1) \textit{Time-series Models}: AMD~\cite{hu2025adaptive}, TimeKAN~\cite{huang2025timekan}, DUET~\cite{qiu2025duet}, LightGTS~\cite{wanglightgts};
        (2) \textit{Spatio-temporal Graph Models}: DCRNN~\cite{li2018diffusion}, STGCN~\cite{yu2018spatio}, GraphWaveNet~\cite{wu2019graph}, ColaGNN~\cite{deng2020cola}, MTGNN~\cite{wu2020connecting}, AGCRN~\cite{bai2020adaptive}, CovidGNN~\cite{kapoor2020examining}, PopNet~\cite{gao2022popnet}, STDDE~\cite{long2024unveiling}.
    \item \textbf{Hybrid Models}: STAN~\cite{gao2021stan}, MPSTAN~\cite{mao2024mpstan}, CausalGNN~\cite{wang2022causalgnn}, MepoGNN~\cite{cao2022mepognn}, EISTGNN~\cite{han2025epidemiology}.
\end{itemize}
Detailed descriptions can be found in Appendix~\ref{sec:appendix_baselines}.

\sstitle{Evaluation Methods}
Following previous work~\cite{tang2023enhancing,cao2022mepognn}, we evaluate the 3d, 7d, 14d, and overall prediction performance
of our model by four metrics, i.e., Root Mean Squared Error (RMSE), Mean Absolute Error (MAE), Symmetric Mean Absolute Percentage Error (SMAPE), and Relative Absolute Error (RAE).

\begin{table*}[t]
\centering
\fontsize{7pt}{8pt}\selectfont
\setlength{\tabcolsep}{3.5pt}
\renewcommand{\arraystretch}{1.15}
\begin{tabular}{lcccccccccccccccc}
\toprule
{\textbf{Model}} 
& \multicolumn{4}{c}{\textbf{3 d Ahead}} 
& \multicolumn{4}{c}{\textbf{7 d Ahead}} 
& \multicolumn{4}{c}{\textbf{14 d Ahead}} 
& \multicolumn{4}{c}{\textbf{Overall}} \\
\cmidrule(lr){2-5}\cmidrule(lr){6-9}\cmidrule(lr){10-13}\cmidrule(lr){14-17}
& RMSE & MAE & SMAPE & RAE
& RMSE & MAE & SMAPE & RAE
& RMSE & MAE & SMAPE & RAE
& RMSE & MAE & SMAPE & RAE \\
\midrule
SIR [1927]       & 148.5 & 92.1 & 193.8 & 1.04
                 & 150.2 & 93.0 & 195.2 & 1.04
                 & 147.3 & 91.0 & 196.8 & 1.05
                 & 149.4 & 92.5 & 195.1 & 1.04 \\
MetaSIR [KDD18]  & 88.4 & 56.5 & 64.2 & 0.64
                 & 124.6 & 80.3 & 80.3 & 0.90
                 & 199.4 & 125.9 & 100.4 & 1.46
                 & 137.3 & 83.9 & 79.9 & 0.95 \\ \hline
STGCN [IJCAI18]  & 62.2 & 29.7 & 32.7 & 0.34
                 & 93.9 & 45.5 & 42.0 & 0.51
                 & 113.7 & 52.4 & 47.6 & 0.61
                 & 92.3 & 42.5 & 40.7 & 0.48 \\
MTGNN [KDD20]    & 135.3 & 81.1  & 64.9 & 0.92
                 & 144.2 & 88.6 & 73.4 & 0.99
                 & 150.5 & 93.6  & 78.7 & 1.08
                 & 143.4 & 87.6  & 71.9 & 0.88 \\
CovidGNN [2020]  & 152.3 & 136.5  & 113.4 & 1.54
                 & 157.5 & 142.2  & 115.5 & 1.60
                 & 167.0 & 152.3  & 119.6 & 1.76
                 & 157.3 & 141.9  & 115.7 & 1.60 \\
DCRNN [ICLR18]   & 57.2 & 28.0 & 32.6 & 0.32
                 & 81.5 & 40.6 & 41.0 & 0.46
                 & 95.8 & 47.0 & 46.7 & 0.54
                 & 80.7 & 38.9 & 40.1 & 0.44 \\
PopNet [WWW22]   & 122.9 & 78.6 & 95.7 & 0.89
                 & 124.0 & 78.8 & 95.3 & 0.88
                 & 121.1 & 76.6 & 95.1 & 0.89
                 & 123.3 & 78.4 & 95.4 & 0.89 \\  
ColaGNN [CIKM20] & 125.2 & 110.5  & 109.4 & 1.25
                 & 127.2 & 112.9  & 110.2 & 1.27
                 & 129.4 & 116.7  & 112.1 & 1.35
                 & 126.7 & 112.1  & 110.0 & 1.27 \\
TimeKAN [ICLR25] & 117.9 & 83.8 & 98.9 & 0.95
                 & 122.6 & 77.3 & 118.6 & 0.87
                 & 121.3 & 84.4 & 99.3 & 0.98
                 & 126.9 & 86.6 & 113.2 & 0.98 \\
COVID-Forecaster [ICDE23]& 119.5 & 69.6 & 87.9 & 0.79
                 & 122.7 & 72.0 & 89.3 & 0.81
                 & 121.7 & 70.7 & 89.8 & 0.82
                 & 122.5 & 71.5 & 89.2 & 0.81 \\
GraphWaveNet [IJCAI19]& 56.8 & 27.7 & 32.4 & 0.31
                 & 80.3 & 39.5 & \underline{40.2} & 0.44
                 & 94.1 & 45.9 & \underline{46.0} & 0.53
                 & 79.1 & 37.7 & \underline{39.3} & 0.43 \\
LightGTS [ICML25]& 94.5 & 53.7 & 71.9 & 0.61
                 & 95.8 & 54.5 & 72.6 & 0.61
                 & 93.4 & 53.0 & 72.2 & 0.61
                 & 95.1 & 54.1 & 72.4 & 0.61 \\
AMD [AAAI25]       & 113.6 & 68.2 & 90.2 & 0.77
                 & 145.3 & 88.2 & 150.1 & 0.99
                 & 129.8 & 74.9 & 106.3 & 0.87
                 & 133.0 & 79.3 & 118.4 & 0.90 \\
STDDE [WWW24]    & 93.8 & 54.8 & 67.3 & 0.62
                 & 96.8 & 57.4 & 70.5 & 0.64
                 & 96.0 & 57.3 & 72.2 & 0.66
                 & 96.2 & 56.9 & 69.8 & 0.64 \\
PDFormer [AAAI23]& 137.6 & 78.5 & 94.3 & 0.89
                 & 137.9 & 78.6 & 94.8 & 0.88
                 & 135.5 & 76.8 & 95.8 & 0.89
                 & 136.2 & 77.6 & 93.6 & 0.88 \\
DUET [KDD25]     & 111.8 & 70.8 & 80.4 & 0.80
                 & 116.6 & 73.9 & 82.6 & 0.83
                 & 147.6 & 91.1 & 183.8 & 1.05
                 & 116.7 & 72.6 & 88.7 & 0.82 \\ 
AGCRN [NeurIPS20]& 139.9 & 81.1  & 101.4 & 0.92
                 & 141.0 & 81.5  & 102.2 & 0.91
                 & 137.7 & 79.1  & 102.2 & 0.92
                 & 140.2 & 81.1  & 102.0 & 0.92 \\ \hline
EISTGNN [EAAI25] & 88.8 & 54.4 & 68.8 & 0.61
                 & 94.5 & 57.9 & 74.3 & 0.65
                 & 95.0 & 58.9 & 77.0 & 0.68
                 & 94.1 & 57.4 & 73.3 & 0.65 \\
STAN [JAMIA21]   & 149.9 & 93.3 & 184.0 & 1.05
                 & 151.0 & 93.7 & 183.6 & 1.05
                 & 147.5 & 91.0 & 183.0 & 1.05
                 & 150.2 & 93.2 & 183.7 & 1.05 \\
CausalGNN [AAAI22] & 121.5 & 78.3 & 96.8 & 0.88
                 & 124.0 & 78.7 & 96.2 & 0.88
                 & 121.6 & 76.6 & 95.5 & 0.89
                 & 123.1 & 78.3 & 96.1 & 0.88 \\
MPSTAN [Entropy24] & 74.3 & 43.1 & 61.0 & 0.49
                 & 83.5 & 47.3 & 63.8 & 0.53
                 & 88.7 & 49.3 & 66.3 & 0.57
                 & 82.6 & 46.7 & 63.5 & 0.53 \\
MepoGNN [ECML-PKDD22] & \underline{54.1} & \underline{26.4} & \underline{32.0} & \underline{0.30}
                 & \underline{72.9} & \underline{37.2} & 40.9 & \underline{0.42}
                 & \underline{84.0} & \underline{43.2} & 46.7 & 0.50
                 & \underline{72.0} & \underline{35.6} & 39.8 & \underline{0.40} \\  \hline \hline
\textbf{STOEP (Ours)}    & \textbf{51.9} & \textbf{24.8} & \textbf{31.6} & \textbf{0.28}
                 & \textbf{65.2} & \textbf{34.4} & \textbf{39.8} & \textbf{0.39}
                 & \textbf{70.5} & \textbf{37.8} & \textbf{45.3} & \textbf{0.44}
                 & \textbf{63.5} & \textbf{32.4} & \textbf{38.8} & \textbf{0.37} \\
\bottomrule
\end{tabular}
\caption{Results on the \textbf{Flu} dataset across different horizons (STOEP's improvements are statistically significant, with $p$-value $< 2 \times 10^{-4}$ across all baselines).}
\label{tab:ZJepi_all_horizons}
\end{table*}

\noindent \sstitle{Implementation \& Training Settings} 
We implement STOEP and baselines using PyTorch on an NVIDIA RTX 5060 GPU.
The model is optimized via Adam with a curriculum learning strategy. Detailed configurations and training settings can be found in Appendix~\ref{sec:appendix_training}.

\subsection{Overall Evaluation}  
Table~\ref{tab:JPepi_all_horizons} summarizes the results of different methods on the COVID-19 dataset against all baselines.
STOEP consistently achieves the best performance across almost all horizons and metrics.
On the overall results, it obtains 10.3\% lower RMSE, 8.9\% lower MAE, and 8.7\% lower RAE than MepoGNN, highlighting a substantial improvement in both accuracy and robustness.
The advantage is particularly evident on the 7-day horizon, where STOEP achieves 15.9\% lower RMSE and 13.3\% lower MAE, demonstrating its superiority in mid-term forecasting.

Table~\ref{tab:ZJepi_all_horizons} summarizes results on the Flu dataset.
STOEP again delivers the best performance across all horizons and metrics.
Compared with MepoGNN on the overall results, it achieves 11.8\% lower RMSE, 9.0\% lower MAE and 7.5\% lower RAE.
On the 7-day horizon, STOEP reduces RMSE and MAE by 10.6\% and 7.5\%, respectively, underscoring stable mid-term benefits in a real operational setting.

The significance test confirms that STOEP's improvements are statistically significant, with a permutation $p$-value $< 2 \times 10^{-4}$ across all baselines.

\subsection{Ablation Study}
We conduct ablation experiments on both datasets to quantify the contribution of each component in STOEP 
by comparing the following variants:
\textit{w/o SPE} keeps the architecture of SPE module but routes its inputs through a $1\times1$ convolution whose output is directly fed into GraphWaveNet for parameter estimation.
\textit{w/o CAL} removes the case-based adjacency intensity.
\textit{CAL-C} uses $k$-shape clustering~\cite{paparrizos2015k} within the CAL module to obtain temporal patterns instead of self-learned patterns.
\textit{w/o FMF} replaces the FMF module with a classic MetaSIR for forecasting.

\begin{figure}[t]
  \begin{minipage}{0.48\textwidth} 
    \includegraphics[width=\linewidth]{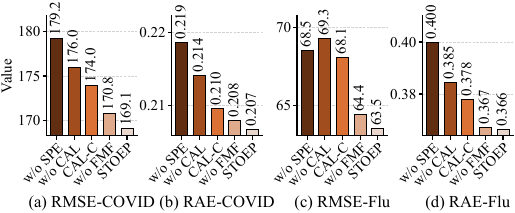}
    \caption{Ablation on the COVID-19 and Flu datasets.}
    \label{fig:Ablation}
  \end{minipage}
  \hfill 
  \begin{minipage}{0.48\textwidth}~\end{minipage} 
\end{figure}

As shown in Figure \ref{fig:Ablation}, 
SPE contributes most, and STOEP lowers RMSE by 5.6\% and RAE by 5.5\% on the COVID-19 dataset, while on the Flu dataset, it lowers RMSE by 7.3\% and RAE by 8.4\%, compared to w/o SPE. 
The CAL module significantly reduced the RMSE by 8.3\% on the Flu dataset. While the CAL-C variant performed better than the w/o CAL variant, it was inferior to STOEP on both datasets. This suggests that the self-learned patterns are more effective than using clustering. 
After the disease transmission suppression in FMF is removed, i.e., w/o FMF, RMSE is increased by 1.4\% on the Flu dataset, which also demonstrates its effectiveness in suppressing unstable parameter estimation.

\begin{figure}[t]
  \begin{minipage}{0.48\textwidth} 
    \includegraphics[width=\linewidth]{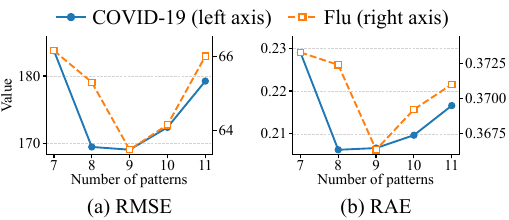}
    \caption{Different number of patterns $P$.}
    \label{fig:DAL_sensitivity}
  \end{minipage}
  \hfill 
  \begin{minipage}{0.48\textwidth}~\end{minipage} 
\end{figure}

\subsection{Hyperparameter Sensitivity}
We further analyzed the sensitivity of the proposed model with respect to the number of learned patterns in the CAL module.
Figure~\ref{fig:DAL_sensitivity} reports the results for RMSE and RAE as the number of patterns varies from 7 to 11 on both the COVID-19 dataset and the Flu dataset.
Overall, the performance is relatively stable across different settings, but performance is generally optimal at 9 patterns.
These results suggest that a moderate number of patterns is sufficient to capture the temporal dynamics without introducing redundancy.

\subsection{Case Study}
\begin{figure}[t]
  \begin{minipage}{0.48\textwidth} 
    \includegraphics[width=\linewidth]{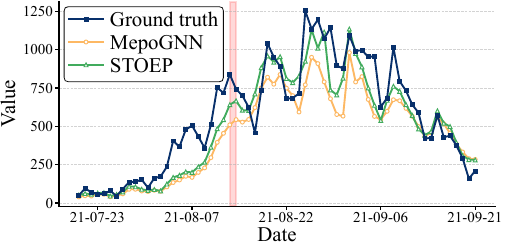}
    \caption{Predicted daily confirmed cases of COVID-19 with horizon = 7.}
    \label{fig:case_study_jp}
    \vspace{-10pt}
  \end{minipage}
  \hfill 
  \begin{minipage}{0.48\textwidth}~\end{minipage} 
\end{figure}

To further demonstrate the effectiveness of STOEP, 
we have conducted a case study on COVID-19 for STOEP and the most competitive forecasting model, i.e., MepoGNN.
We visualize the prediction results (7d ahead) of STOEP and MepoGNN from July 18, 2021, to September 19, 2021, on COVID-19 dataset in Figure~\ref{fig:case_study_jp},
where the blue line represents the ground-truth. It can be observed that our model is closer to the ground-truth than MepoGNN and more capable of capturing the occurrence of peaks. For instance, STOEP successfully predicted a peak around August 13-14, a period when Delta variant was prevalent in Japan and the epidemic was expanding rapidly\footnote{\url{https://japan.kantei.go.jp/99_suga/statement/202108/_00009.html}}. This indicates that our model can reflect real-world scenarios more accurately than MepoGNN. 

The case study on the Flu dataset is also conducted which demonstrates STOEP's ability to capture seasonal peaks and delayed propagation. Details refer to Appendix~\ref{sec:appendix_flu_beta}.

\subsection{System Deployment}

\begin{figure}[t]
  \centering
  \includegraphics[width=\linewidth]{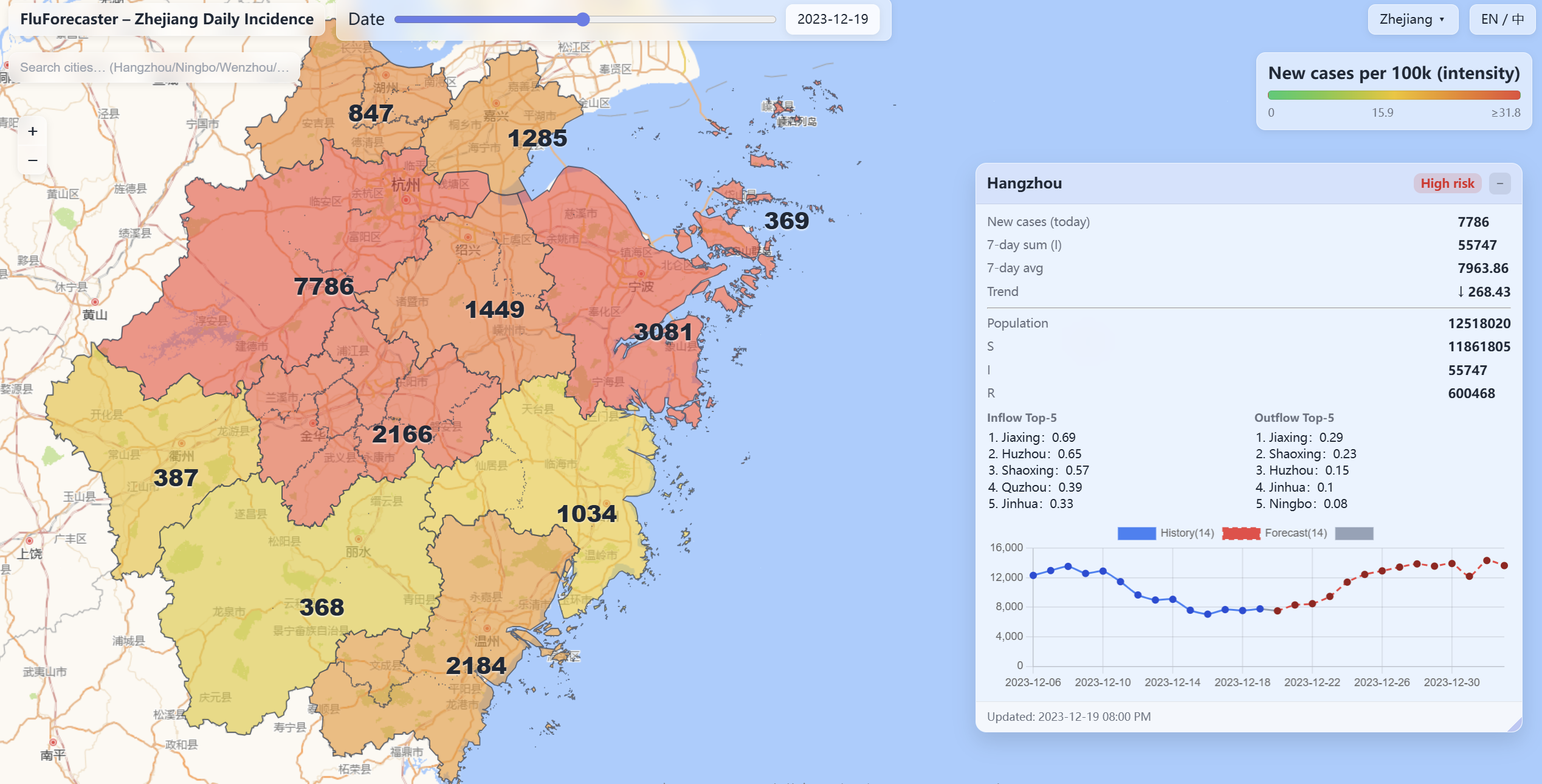}
  \caption{Screenshot of the deployed system based on STOEP at Zhejiang CDC in China.}
  \label{fig:system_interface}
\end{figure}

A system based on STOEP has already been deployed at Zhejiang CDC in China, providing daily influenza forecasts for 11 cities. As shown in Figure~\ref{fig:system_interface}, the system utilizes our model to visualize real-time trends and risk levels.
Public health practitioners can use the system to trigger the early warning of infectious diseases and precisely allocate medical resources among regions in advance.

\section{Related Work}\label{sec:related}

Epidemic forecasting can be mainly categorized into three types: mechanistic models, data-driven models, and hybrid models.

\sstitle{Mechanistic Models}
Mechanistic models model the infecting dynamics across different groups of populations, i.e., compartments, using differential equations, e.g., SIR model~\cite{kermack1927contribution,rizoiu2018sir}, SEIR model~\cite{wang2024precise}. Considering the spatial heterogeneity, MetaSIR~\cite{wang2018inferring} is further proposed to jointly forecast for several regions. However, they usually rely on experts' knowledge and strong assumptions on disease transmission, which limit their prediction capability. 

\sstitle{Data-driven Models}
Data-driven models can flexibly learn to forecast based on past observations, which are widely studied in epidemic forecasting and other domains, e.g., traffic forecasting. Auto-Regressive model~\cite{perrotta2017using}, ARIMA~\cite{kufel2020arima} are initially adopted. 
In the era of deep learning, Bi-LSTM~\cite{arora2020prediction} is commonly used. For spatio-temporal forecasting, the spatial dependency among different regions can be constructed based on region adjacency~\cite{deng2020cola,kapoor2020examining}, mobility intensity~\cite{gao2022popnet}, or self-attention~\cite{shen2023forecasting,zou2026meta}. Though data-driven models show superior performance, they learn parameters purely based on data, which limits their generalization capability.

\sstitle{Hybrid Models}
Hybrid models combine the mechanistic models and powerful modeling ability of deep learning. STAN~\cite{gao2021stan} and PISID~\cite{fujita2025enhancing} regulate the network training loss with SIR constraints, while MPSTAN~\cite{mao2024mpstan} adds the MetaSIR module into the recurrent cell. 
MepoGNN~\cite{cao2022mepognn}, CausalGNN~\cite{wang2022causalgnn} and EISTGNN~\cite{han2025epidemiology} estimate epidemic parameters of mechanistic models by neural networks. 
The hybrid idea is also used in other domains, for example, CTENet~\cite{zhang2026eulerian} fuses the physical and
chemical terms into the decoder for air quality forecasting, while GravityFormer~\cite{wang2026gravity} fuses the gravity model for human activity intensity prediction.
Our proposed method also belongs to this category. However, different from previous work, we enhance the models with spatio-temporal priors via CAL and SPE, and expert priors via FMF.

\section{Conclusion}

In this paper, we propose STOEP, a spatio-temporal prior-aware epidemic predictor, which enhances the hybrid model by prior knowledge. Spatio-temporal priors are injected into the adjacency learning and amplify the input signals, while expert priors guide the design of parameter adaptive thresholding. Experiments on COVID-19 and influenza datasets show that STOEP consistently outperforms the best baseline by 11.1\% on average in RMSE. A system based on STOEP is also deployed at Zhejiang CDC in China for forecasting to facilitate downstream applications.
Further investigation into the cross-regional and cross-disease transferability of the proposed model is warranted.

% \newpage

\section*{Acknowledgements}
This work was supported by the National Key R\&D Program of China (No. 2023YFC2308703), the National Natural Science Foundation of China (No. 62306033, 42371480).

\bibliographystyle{named}
\bibliography{ijcai26}

\clearpage
\appendix

\section{Complexity and Scalability Analysis}
\label{sec:appendix_complexity}
We analyze the computational complexity of STOEP to assess its efficiency and scalability.
Let $N$ denote the number of regions, $T_{\text{in}}$ the input sequence length, $T_{\text{out}}$ the forecasting horizon, and $C$ the feature dimension. Batch size and other constant factors are omitted for clarity.

\paragraph{Time Complexity.}
The dominant computational cost of STOEP arises from dense region-to-region interactions. The dynamic adjacency learning and spatial modeling operations involve $N \times N$ matrix computations, while spatial attention and graph convolution are applied across the temporal dimension. As a result, the overall time complexity of STOEP is
\begin{equation}
\mathcal{O}\left(N^2 \left(T_{\text{in}} C + T_{\text{out}}\right)\right).
\end{equation}

\paragraph{Space Complexity.}
STOEP requires memory to store dense adjacency matrices, mobility-related origin--destination tensors, and intermediate node representations. Specifically, the space complexity is dominated by the storage of $N \times N$ matrices across both input and prediction horizons, leading to
\begin{equation}
\mathcal{O}\left((T_{\text{in}} + T_{\text{out}}) N^2 + T_{\text{in}} N C\right)
\end{equation}
memory usage.

\paragraph{Scalability.}
Although STOEP exhibits a quadratic dependency on the number of regions, this design is well suited for epidemic forecasting tasks at the city or prefecture level, where the number of regions typically ranges from tens to a few hundreds, making the model computationally practical in real-world settings.

\section{Dataset Characteristics and Generalizability Analysis}
\label{sec:appendix_dataset_analysis}

To thoroughly assess the robustness of STOEP against demographic bias and diverse epidemic dynamics, we analyze the statistical properties of the two datasets used in our evaluation.

\begin{figure}[htbp]
  \centering
  \includegraphics[width=0.8\linewidth]{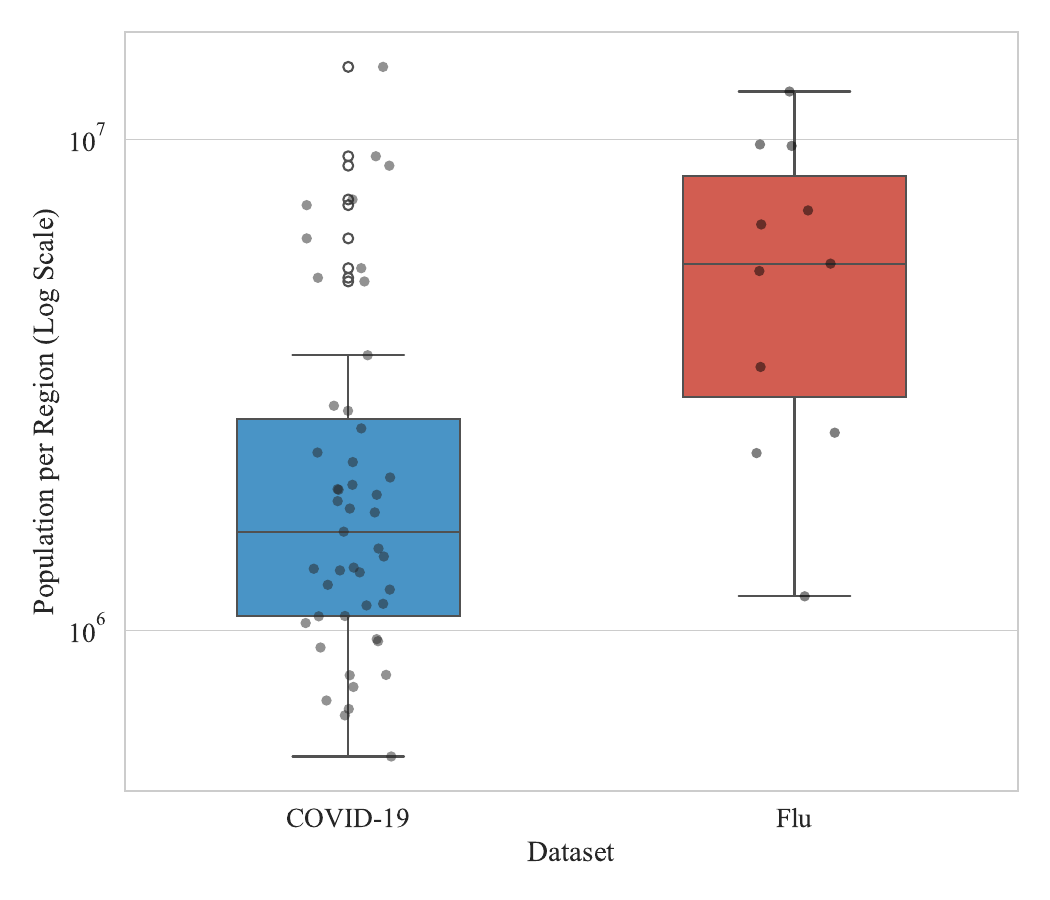}
  \caption{Comparison of regional population distributions (Log Scale). The COVID-19 dataset covers a national scale with high variance, while the Flu dataset represents a provincial scale.}
  \label{fig:demographic_comparison}
\end{figure}

\paragraph{Demographic Scale.}
As illustrated in Figure~\ref{fig:demographic_comparison}, the two datasets represent distinct demographic scales. The Japan dataset covers a national scale ($N=47$ prefectures) with a high variance in population sizes, ranging from major metropolitan areas to rural prefectures. In contrast, the Zhejiang dataset focuses on a provincial scale ($N=11$ cities) characterized by relatively denser and more uniform urban populations. This difference allows us to validate STOEP's performance across varying administrative granularities.

\begin{figure}[htbp]
  \centering
  \includegraphics[width=0.9\linewidth]{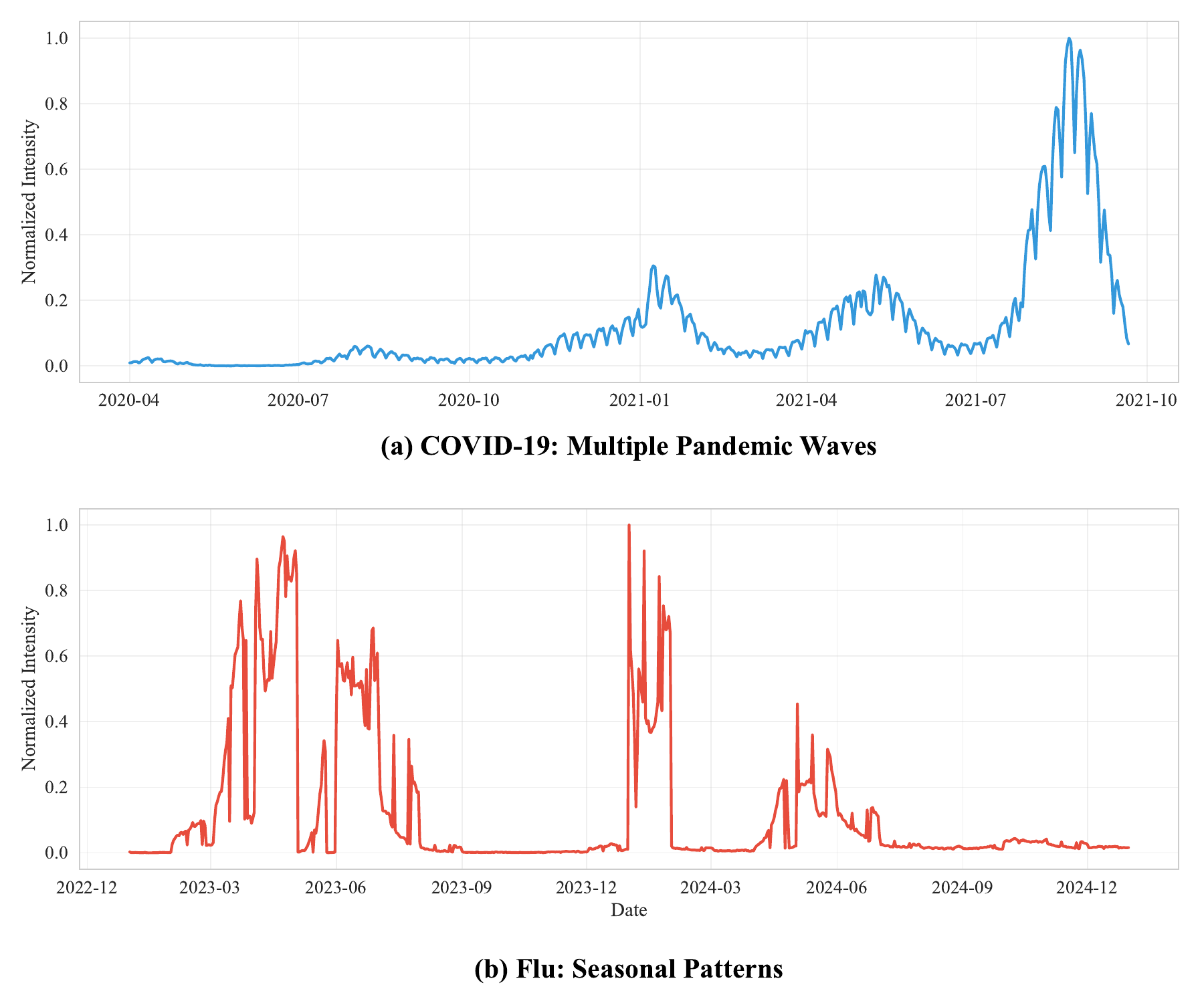}
  \caption{Normalized daily confirmed cases. (Top) COVID-19 in Japan exhibits multiple pandemic waves driven by variants. (Bottom) Flu in China shows characteristic seasonal peaks.}
  \label{fig:epidemic_dynamics}
\end{figure}

\paragraph{Epidemic Dynamics.}
Figure~\ref{fig:epidemic_dynamics} visualizes the normalized daily confirmed cases to highlight the temporal patterns of the diseases while preserving data privacy. The COVID-19 dataset exhibits complex, irregular pandemic waves driven by the emergence of new variants. Conversely, the Flu dataset demonstrates distinct seasonal peaks typical of influenza. STOEP's superior performance on both datasets confirms its capability to generalize across biologically distinct pathogens.

\begin{figure}[htbp]
  \centering
  \includegraphics[width=0.8\linewidth]{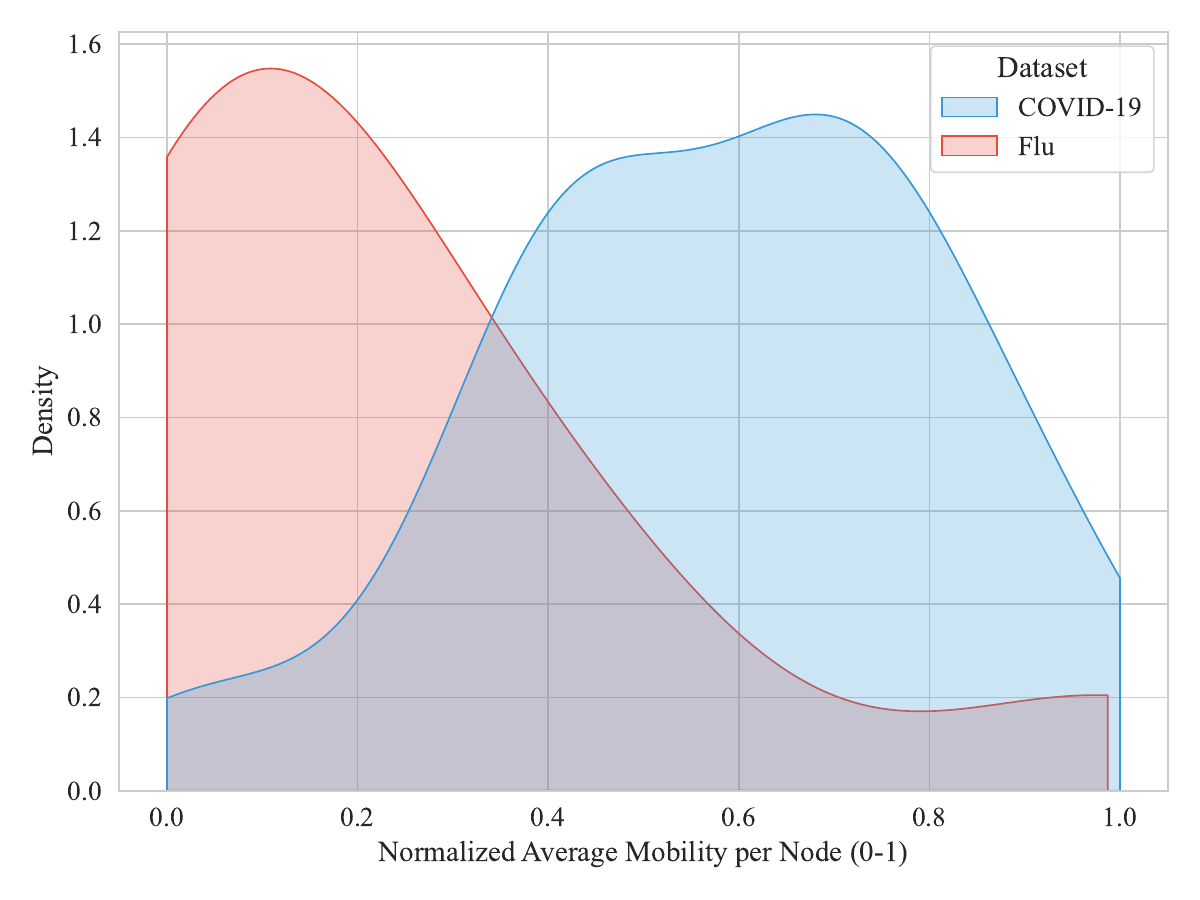}
  \caption{Distribution of normalized average mobility intensity per node. Despite different sources (Facebook vs. Baidu), both datasets provide rich mobility heterogeneity.}
  \label{fig:mobility_distribution}
\end{figure}

\paragraph{Mobility Distribution.}
We also analyze the mobility data characteristics in Figure~\ref{fig:mobility_distribution}. The COVID-19 dataset utilizes Facebook Movement Range Maps, while the Flu dataset uses Baidu Migration Map data. Despite originating from different providers with different raw scales, the normalized distribution shows that both datasets provide rich heterogeneity in mobility intensity. STOEP effectively leverages these diverse mobility signals to enhance adjacency learning.

In summary, by demonstrating consistent improvements across diverse settings, encompassing national and provincial demographics, pandemic and seasonal disease dynamics, and distinct mobility data sources, STOEP evidences strong generalizability and robustness against potential demographic biases.

% --- 新增部分：详细的基线模型介绍 ---
\section{Detailed Baseline Descriptions}
\label{sec:appendix_baselines}
We compare our proposed method with three types of baselines as mentioned in Section~\ref{sec:related}: 
1) Mechanistic Models, 2) Deep Learning Models, and 3) Hybrid Models. We detail them as follows:

\begin{enumerate}[leftmargin=*]
    \item \textbf{Mechanistic Models}. 
    We select two representative methods, i.e., SIR~\cite{kermack1927contribution} and MetaSIR~\cite{wang2018inferring}.
    \item \textbf{Deep Learning Models}. 
    We compare the proposed method with epidemic-specific deep learning models as well as general time-series or spatio-temporal models, which are further categorized as follows:
        \begin{itemize}[leftmargin=*]
            \item Time-series Models, which include AMD~\cite{hu2025adaptive}, TimeKAN~\cite{huang2025timekan}, DUET~\cite{qiu2025duet}, LightGTS~\cite{wanglightgts}. To adapt to the spatio-temporal epidemic forecasting problem, we transform the historical observations into a matrix by flattening the region and factor dimensions. Besides, to further integrate the mobility data, we flatten, embed and concatenate it with other observations. The model outputs are transformed back to the region dimension to obtain the final predictions.
            \item Spatio-temporal Graph Models, which construct a spatial graph to model spatial dependency, including 
            DCRNN~\cite{li2018diffusion}, STGCN~\cite{yu2018spatio}, GraphWaveNet~\cite{wu2019graph}, 
            ColaGNN~\cite{deng2020cola},
            MTGNN~\cite{wu2020connecting}, 
            AGCRN~\cite{bai2020adaptive}, 
            CovidGNN~\cite{kapoor2020examining},
            PopNet~\cite{gao2022popnet} and STDDE~\cite{long2024unveiling}. Among them, only CovidGNN~\cite{kapoor2020examining} leverages the mobility data to set the edge weights. 
            For DCRNN~\cite{li2018diffusion}, STGCN~\cite{yu2018spatio}, PopNet~\cite{gao2022popnet}, and STDDE~\cite{long2024unveiling}, we follow a common practice by constructing a static spatial graph where the edge weights are derived from average inter-region mobility flows. For other graph models that learn dynamic adjacency matrices, the time-varying mobility graph is integrated through temporal weighted fusion with the learned dynamic graph.
        \end{itemize}
    \item \textbf{Hybrid Models.} 
    We compare network-dominated forecasters, i.e., STAN~\cite{gao2021stan}, MPSTAN~\cite{mao2024mpstan}, as well as mechanism-dominated forecasters, i.e., CausalGNN~\cite{wang2022causalgnn}, MepoGNN~\cite{cao2022mepognn}.
    Among them, MepoGNN~\cite{cao2022mepognn} leverages the mobility data to derive the edge weights.
    For CausalGNN~\cite{wang2022causalgnn}, mobility information is incorporated by combining the attention-based dynamic graph and the time-varying mobility graph through weighted fusion. 
    For EISTGNN~\cite{han2025epidemiology}, we use time-varying mobility flows to represent dynamic inter-regional interactions and their temporal average to construct a static mobility support.
    For other hybrid models, we follow a standard practice by constructing a static adjacency matrix where edge weights are derived from aggregated mobility data.
\end{enumerate}

\section{Implementation and Training Details}
\label{sec:appendix_training}

The input time length $T_{\text{in}}$ and output time length $T_{\text{out}}$ are both set to 14 days, 
which means we use two-week historical observations to do the two-week prediction of daily confirmed cases.
$P$ is set to 9. 
$H$ is set to 4.
$\kappa_I$, $\kappa_\rho$, $\kappa_\beta$ and $\kappa_\gamma$ are set to 0.1, 0.9, 0.2 and 0.2.
$\eta$ is 0.9.
$\delta_{\min}^I$, $\delta_{\min}^\rho$, $\delta_{\min}^\beta$ and $\delta_{\min}^\gamma$ are set to 0.5, 0.7, 2e-3 and 2e-3.
$\kappa_{\rho}^{\max}$ is set to 0.98.
$\psi$ is set to 0.5.

During the training, the batch size is set to 32. Following previous work~\cite{cao2022mepognn}, we use the curriculum learning strategy~\cite{wu2020connecting} to gradually increase the prediction horizon.
Adam~\cite{kingma2014adam} optimizer is used, where the learning rate is $1\text{e-3}$ and weight decay is $1\text{e-8}$.
The training algorithm would either be early-stopped if the validation error did not decrease within 20 epochs or be stopped after 300 epochs.

We implement our model as well as baselines with Python and PyTorch, and train them on a server with an NVIDIA RTX 5060 GPU.

\begin{figure}[b]
  \begin{minipage}{0.48\textwidth} % 左半页
    % 去掉 \centering 就不会居中
    \includegraphics[width=\linewidth]{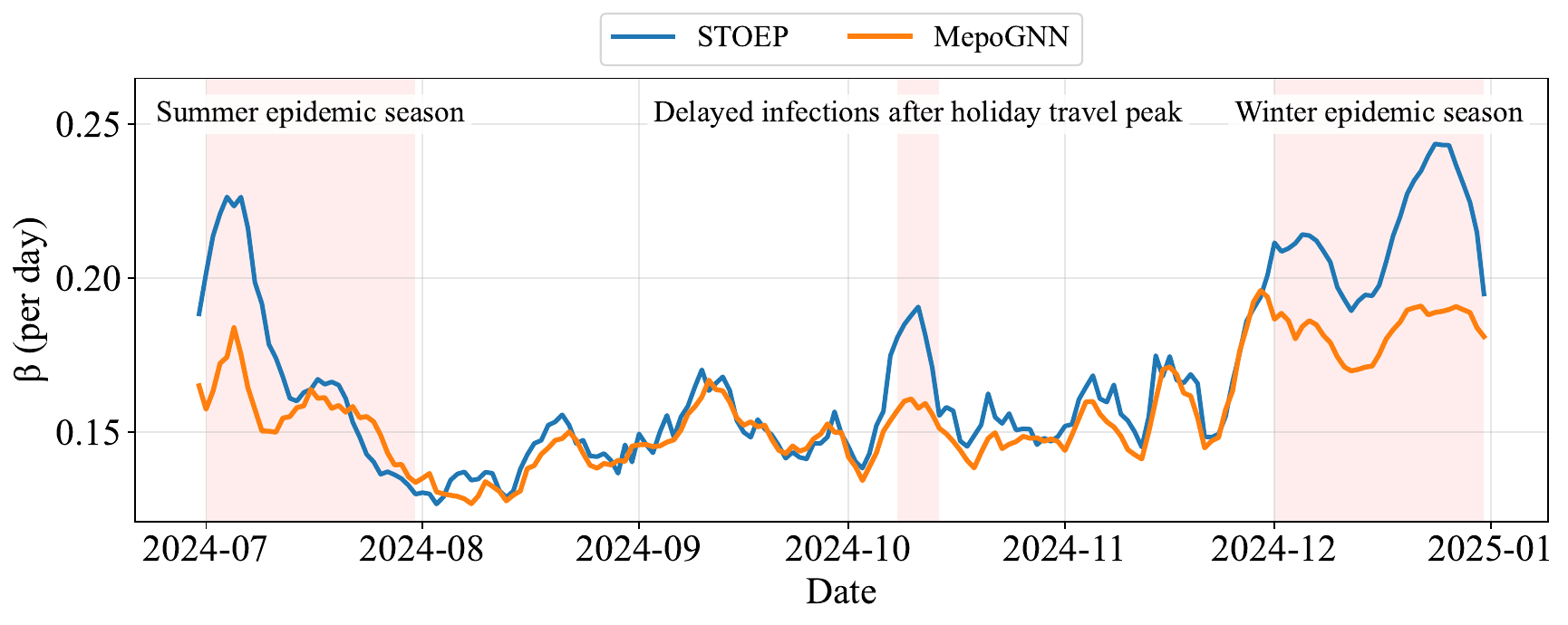}
    \caption{7-day moving average of predicted $\beta$ of Flu with horizon = 7.}
    \label{fig:case_study_beta_hz}
  \end{minipage}
  \hfill % 中间留一点弹性空白
  \begin{minipage}{0.48\textwidth}~\end{minipage} % 右半留空
\end{figure}

\section{Case Study on Flu Epidemic Parameters}
\label{sec:appendix_flu_beta}
We analyze the estimated epidemiological parameters to demonstrate the model's interpretability.
Figure~\ref{fig:case_study_beta_hz} presents estimated $\beta$ from July 2024 to December 2024 in the Flu dataset, where a larger $\beta$ indicates that more susceptible individuals become infected. We can observe that STOEP exhibits three more distinct peaks than MepoGNN, indicating that STOEP is more capable of capturing the summer and winter epidemic season, and successfully captures a delayed epidemic peak induced by holiday travel in October, indicating that our model is better able to reflect infection peak periods.

\section{Metapopulation SIR}~\label{sec:metasir}

Metapopulation SIR model, i.e., MetaSIR~\cite{wang2018inferring} extends classic SIR model~\cite{kermack1927contribution} by assuming the heterogeneity of regions and using human mobility to model the propagation among regions.
For each region $n$, the mobility-induced transmission strength of region $n$
at time $t+1$, i.e., $\Pi_n^{t+1}$, is defined as:
\begin{align}
\Pi_n^{t+1} = \sum_{m=1}^N \Bigg(\frac{h_{nm}^{t+1}}{P_m} + \frac{h_{mn}^{t+1}}{P_n}\Bigg) I_m^{t}, 
\label{eq:transmission}
\end{align}
where $h_{nm}^{t+1}$ denotes the mobility flow from region $n$ to $m$, $P_m$ is the population size of region $m$, $I_m^{t}$ is the infected population in region $m$. The epidemic dynamics for region $n$ is as follows:
\begin{equation}~\label{eq:epi_dyn}
\begin{aligned}
\frac{dS_n^{t+1}}{dt} &= -\beta_n^{t+1}\Pi_n^{t+1}\\
\frac{dI_n^{t+1}}{dt} &= \beta_n^{t+1}\Pi_n^{t+1} - \gamma_n^{t+1}I_n^{t} \\
\frac{dR_n^{t+1}}{dt} &= \gamma_n^{t+1} I_n^{t}
\end{aligned}
\end{equation}
where $\beta_n^{t+1}$ and $\gamma_n^{t+1}$ are the infection and recovery parameters of region $n$, and $S_n^{t+1}, I_n^{t+1}, R_n^{t+1}$ denote the susceptible, infected, and recovered populations of region $n$ at day $t+1$. 
Based on the epidemic dynamics in Equation~\ref{eq:epi_dyn}, $S_n^{t+1}, I_n^{t+1}, R_n^{t+1}$ can be derived via differential equations, and the predicted daily confirmed cases of region $n$ at day $t+1$ can be calculated as:
\begin{equation}~\label{eq:pred}
    \hat{x}^{t+1}_n=\beta^{t+1}_n \Pi_n^{t+1}
\end{equation}
When this process is called recursively, we can obtain the multi-step daily confirmed case predictions $\hat{\mathbf{X}}^{t+1:t+T_{out}}$:
\begin{equation}
    \hat{\mathbf{X}}^{t+1:t+T_{out}}=
    \begin{bmatrix}
    \hat{x}^{t+1}_1 & \dots  & \hat{x}^{t+T_{out}}_1 \\
    \vdots & \ddots & \vdots \\
    \hat{x}^{t+1}_N & \dots  & \hat{x}^{t+T_{out}}_N
\end{bmatrix}
\end{equation}

Classical SIR treats each region in isolation and relies on fixed-form infection and recovery dynamics, which limits its ability to capture mobility-driven transmission and to handle time-varying parameters robustly. The epidemic dynamics for region $n$ follow:
\begin{align}
\frac{dS_n^{t+1}}{dt} &= -\beta_n^{t+1}I_n^{t}, \label{eq:sir_s}\\
\frac{dI_n^{t+1}}{dt} &= \beta_n^{t+1}I_n^{t} - \gamma_n^{t+1}I_n^{t}, \label{eq:sir_i}\\
\frac{dR_n^{t+1}}{dt} &= \gamma_n^{t+1} I_n^{t}, \label{eq:sir_r}
\end{align}
where $\beta_n^{t+1}$ and $\gamma_n^{t+1}$ are the infection and recovery parameters of node $n$.

\end{document}